\documentclass[sigconf,screen]{acmart}
\renewcommand\footnotetextcopyrightpermission[1]{}
\usepackage{booktabs}
\usepackage{multicol}
\usepackage{multirow}

\usepackage{amsmath,amssymb,amsfonts}
\usepackage{algorithmic}
\usepackage{graphicx}
\usepackage{textcomp}
\usepackage{xcolor}
\usepackage{subcaption}
\usepackage{enumitem}
\AtBeginDocument{%
  }

\acmConference[DAC '2026]{The Chips to Systems Conference}{July 26--29,
  2026}{Long Beach, CA}
  
\begin{document}

\title{EARL: Entropy-Aware RL Alignment of LLMs for Reliable RTL Code Generation}



\vspace{20 pt}
\author{Jiahe Shi}
\affiliation{%
  \institution{MIT}
  \country{}
}
\author{Zhengqi Gao}
\affiliation{%
  \institution{MIT}
  \country{}
}
\author{Ching-Yun Ko}
\affiliation{%
  \institution{IBM}
  \country{}
}
\author{Duane Boning}
\affiliation{%
  \institution{MIT}
  \country{}
}
\vspace{20 pt}

\renewcommand{\shortauthors}{Trovato et al.}

\begin{abstract}
Recent advances in large language models (LLMs) have demonstrated significant potential in hardware design automation, particularly in using natural language to synthesize Register-Transfer Level (RTL) code. 
Despite this progress, a gap remains between model capability and the demands of real-world RTL design, including syntax errors,
functional hallucinations, and weak alignment to designer intent.
Reinforcement Learning with Verifiable Rewards (RLVR) offers a promising approach to bridge this gap, as hardware provides executable and formally checkable signals that can be used to further align model outputs with design intent.
However, in long, structured RTL code sequences, not all tokens contribute equally to functional correctness, and naïvely spreading gradients across all tokens dilutes learning signals.
A key insight from our entropy analysis in RTL generation is that only a small fraction of tokens (e.g., \texttt{always}, \texttt{if}, \texttt{assign}, \texttt{posedge}) exhibit high uncertainty and largely influence control flow and module structure.
To address these challenges, we present EARL, an Entropy-Aware Reinforcement Learning framework for Verilog generation. 
EARL performs policy optimization using verifiable reward signals and introduces entropy-guided selective updates that gate policy gradients to high-entropy tokens.
This approach preserves training stability and concentrates gradient updates on functionally important regions of code.
Our experiments on VerilogEval and RTLLM show that EARL improves functional pass rates over prior LLM baselines by up to 14.7\%, while reducing unnecessary updates and improving training stability. These results indicate that focusing RL on critical, high-uncertainty tokens enables more reliable and targeted policy improvement for structured RTL code generation.
We will release the code upon acceptance. An anonymized repository for review is available at https://anonymous.4open.science/r/EARL-1C25.
\end{abstract}

\maketitle

\section{Introduction}

Large Language Models (LLMs) have achieved remarkable success in natural language processing and general-purpose code generation, demonstrating strong capabilities in understanding context, modeling semantics, and synthesizing coherent outputs across a wide range of tasks~\cite{2024llmse_survey,2025llmcode_survey}. While these models have revolutionized software development workflows, applying LLMs to specialized domains such as hardware design remains a challenging frontier~\cite{2025llmeda_survey}.

One such challenge is the generation of Register-Transfer Level (RTL) code, typically written in Verilog. RTL code lies as the foundation for digital hardware design, translating architectural intent into synthesizable logic. 
Writing RTL code requires deep domain knowledge and significant engineering effort. Small errors in syntax, port declarations, or timing semantics can lead to non-compilable or functionally incorrect designs. 
Automating RTL code generation has the potential to significantly accelerate hardware development cycles, reduce engineering effort, and make hardware design more accessible. 
Recent efforts have explored applying LLMs to Verilog generation, with promising results in syntactic fluency and basic design translation~\cite{nyu2022benchmark,chipnemo,2023chipgpt}.


However, achieving reliable RTL generation remains challenging.
Most existing approaches rely on supervised fine-tuning (SFT) on pre-collected datasets~\cite{pku2024origen,xie2024rtlcoder}. While this improves syntactic fluency, compiler and testbench feedback, which is essential for functional correctness, remains unused during training. 
This motivates reinforcement learning with verifiable rewards (RLVR)~\cite{2025rl_verilog}, which leverages executable and formally checkable signals to align model outputs with design intent. 
Yet, applying RLVR to RTL code generation introduces two fundamental difficulties. 
First, compiler and testbench feedback is inherently sparse and delayed, making credit assignment particularly challenging in long, structured code sequences~\cite{2025beyond}. 
Second, standard RLVR methods spread updates uniformly across all tokens, even though only a small subset governs structural and functional correctness. 


\begin{figure}[tb]
\centerline{
\includegraphics[width=0.85\linewidth]{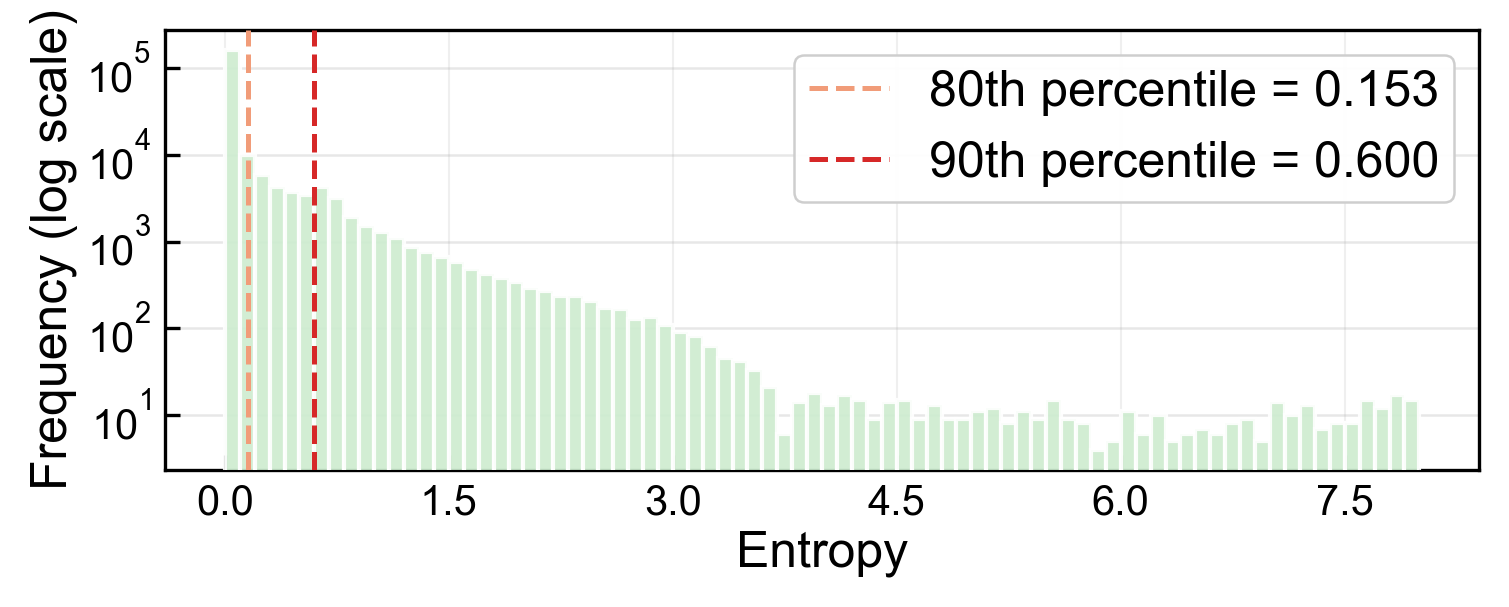}
}
\vspace{-15pt}
\caption{Token-level entropy distribution in RTL generation.}
\label{fig:entropy_hist}
\vspace{-15pt}
\end{figure}

To understand where learning signals should be allocated, we analyze token-level entropy in RTL generation.
As shown in Figure~\ref{fig:entropy_hist}, the entropy distribution is highly skewed. Most tokens are generated with near-zero entropy, while only a small fraction exhibit significantly higher uncertainty. 
This pattern echoes recent findings in RLVR research~\cite{2025entropy_review,2025beyond,2025stabilizing}.
In RTL, the imbalance is even more pronounced, since deterministic syntax and port declarations dominate sequences.
These observations suggest that uniform updates waste gradient budget, motivating reinforcement learning (RL) methods that selectively focus on high-entropy, high-impact positions. A more detailed entropy study is presented in Section \ref{sec:entropy}.


To address these challenges, we propose \textbf{EARL}, an \emph{Entropy-Aware Reinforcement Learning} framework for Verilog generation.
EARL combines supervised initialization with verifiable multi-signal rewards (syntax, interface, functionality) and introduces entropy-guided selective updates.
By concentrating policy gradients on uncertain and functionally critical tokens, EARL improves alignment with hardware correctness while preserving training stability.
Our contributions are summarized as follows:
\begin{enumerate}[leftmargin=16pt]
    \item \textbf{Token-level entropy analysis for RTL.} 
    We provide the first entropy study in RTL generation.
    Our analysis shows that while most tokens are deterministic, a small fraction of high-entropy tokens significantly contribute to functional correctness.
    This motivates the need for reinforcement learning methods that update policies selectively rather than uniformly.
    
    \item \textbf{Entropy-aware reinforcement Learning framework.} 
    We introduce EARL, an entropy-aware RL framework that integrates verifiable rewards with selective policy updates on high-entropy tokens.
    Unlike uniform update schemes, EARL targets high-entropy tokens while preserving coherence on low-entropy ones, improving alignment with hardware correctness and stabilizing training dynamics.
    The design is optimizer-agnostic and can be layered on standard objectives. We instantiate it with both Proximal Policy Optimization (PPO) and Dynamic Sampling Policy Optimization (DAPO) in this work.
    
    \item \textbf{Verifiable multi-signal reward aligned with RTL practice.} 
    We combine compiler validity, interface consistency, and testbench outcomes into a verifiable reward that reflects hardware design constraints. 
    This multi-signal feedback provides executable and formally checkable guidance that goes beyond textual fluency, aligning generation with downstream RTL verification requirements. 

    \item \textbf{Verilog code generation performance improvement.} 
    EARL consistently improves the functional correctness of LLMs on standard RTL benchmarks. 
    Without relying on task-specific prompting or iterative repair, EARL surpasses strong supervised and RL baselines, achieving up to \textbf{14.7\%} increase in pass@5 on \textsc{RTLLM}.

\end{enumerate}

\section{Background}

\subsection{LLM for Verilog Code Generation}
Recent progress in LLMs has spurred research into thier application to Verilog code generation~\cite{nyu2022benchmark,xie2024rtlcoder,nyu2024,2024autovcoder,ren2024rtlfixer,2024assertllm,pku2024origen,2024betterv,haven2025}. Existing approaches can be grouped into three categories: \emph{supervised fine-tuning}, \emph{prompt engineering}, and \emph{reinforcement learning}. 

{Supervised fine-tuning (SFT)} methods adapt pre-trained LLMs using hardware-specific datasets. Examples include RTLCoder~\cite{xie2024rtlcoder} and VeriGen~\cite{nyu2024}, which fine-tune on automatically generated instruction–code pairs. Other works enrich data through domain-specific transformations or augmentations. For instance, BetterV~\cite{2024betterv} leverages Verilog-to-C translation. Despite their gains, SFT methods are constrained by limited and noisy datasets and remain static, lacking adaptive feedback.

{Prompt engineering} approaches guide models via specialized prompting or interactive tool calls. 
RTLFixer~\cite{ren2024rtlfixer} employs ReAct prompting and retrieval to iteratively fix syntax errors.
OriGen~\cite{pku2024origen} combines exemplar prompting with self-reflection and code-to-code transformations, while HaVen~\cite{haven2025} introduces a hallucination taxonomy and employs chain-of-thought prompting over symbolic modalities. 
Prompt-based methods can improve syntactic and functional correctness, but often suffer from high computational cost and poor scalability due to reliance on repeated synthesis or tool feedback.

{Reinforcement learning} has recently been explored to bridge the gap between fluency and functional correctness. Building on RLHF~\cite{2020rlhf,2022rlhf}, methods like PPOCoder~\cite{2023ppocoder} and PLUM~\cite{2024plum} leverage compiler or test outcomes as reward signals. For Verilog generation, RLVR-style methods have emerged~\cite{2025rl_1,rl2025verireason}, but existing designs are limited: rewards based on AST similarity can be gamed without functional benefit~\cite{2025rl_1}, while structured reasoning with distillation may constrain diversity~\cite{rl2025verireason}. These limitations highlight the need for reinforcement learning frameworks that leverage verifiable multi-signal feedback while avoiding uniform or inefficient token-level updates.

\begin{figure*}[htp!]
    \centering
    \begin{subfigure}[t]{0.48\textwidth}
        \centering
        \includegraphics[width=\linewidth]{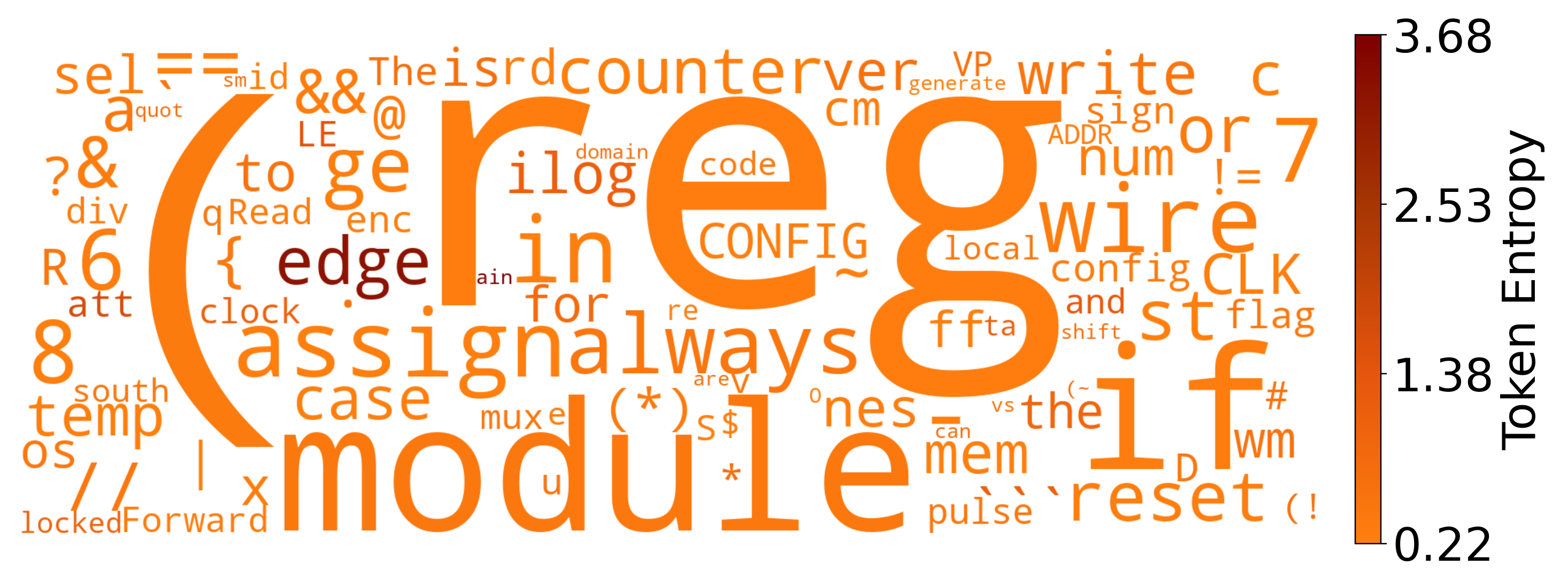}
        \vspace{-20pt}
        \caption{High-entropy tokens}
        \label{fig:wordcloud_high}
    \end{subfigure}
    \hfill
    \begin{subfigure}[t]{0.48\textwidth}
        \centering
        \includegraphics[width=\linewidth]{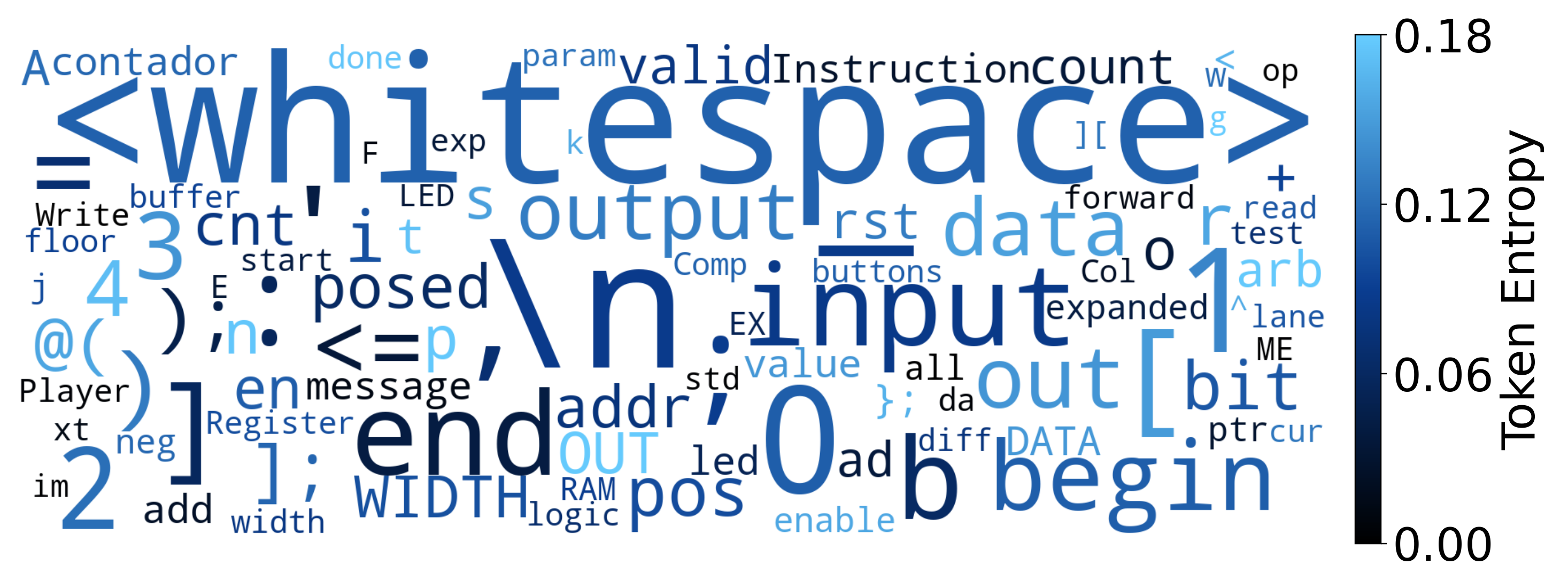}
        \vspace{-20pt}
        \caption{Low-entropy tokens}
        \label{fig:wordcloud_low}
    \end{subfigure}
    \vspace{-10pt}
    \caption{Word cloud visualization of token-level entropy in RTL generation.
    Font size indicates the frequency of a lexical token across the corpus, while color encodes its average entropy (blue = low, orange = high).
    }
    \vspace{-8pt}
    \label{fig:wordcloud}
\end{figure*}

\subsection{RLVR Algorithms} 

\textbf{Proximal Policy Optimization (PPO).}  
PPO~\cite{schulman2017proximal} is a widely used policy gradient algorithm that stabilizes reinforcement learning by clipping the probability ratio between new policy $\pi_{\theta}$ and old policy $\pi_{\theta_{\mathrm{old}}}$. The objective is as follows:

\begin{equation}
\label{eq:ppo}
\begin{aligned}
J_{\mathrm{PPO}}(\theta)
= ~&\mathbb{E}_{(q,a)\sim\mathcal{D},\, \tilde{o}\sim\pi_{\theta_{\mathrm{old}}}(\cdot \mid q)}
\Big[
\min\!\big(
r_t(\theta)\,\hat{A}_t,\, \\
&\operatorname{clip}\!\big(r_t(\theta),\, 1-\epsilon,\, 1+\epsilon\big)\,\hat{A}_t
\big)
\Big],\\
\text{with}\qquad
&r_t(\theta)
= \frac{\pi_{\theta}\!\left(o_t \mid q,\, o_{<t}\right)}
{\pi_{\theta_{\mathrm{old}}}\!\left(o_t \mid q,\, o_{<t}\right)}\, ,
\end{aligned}
\end{equation}
where $\mathcal{D}$ is a dataset of queries $q$ and corresponding ground-truth answers $a$. 
$\tilde{o} = (o_1, \ldots, o_T)$ denotes a sampled output sequence from the old policy $\pi_{\theta_{\mathrm{old}}}$, $\epsilon \in \mathbb{R}$ is a clipping hyperparameter, and $\hat{A}_t$ is the advantage estimate from a learned value function.

\textbf{Group Relative Policy Optimization (GRPO).}  
GRPO~\cite{grpo2024} removes the value function and computes the advantages by normalizing scalar rewards within a group of rollouts sampled from the same prompt. This group-wise normalization removes the need for a critic, reducing memory cost and improving stability.
The objective is as follows:
\begin{equation}
\label{eq:grpo}
\begin{aligned}
J_{\mathrm{GRPO}}(\theta)
=&~ \mathbb{E}_{(q,a)\sim\mathcal{D},\, \{\tilde{o}^{i}\}_{i=1}^{G}\sim
\pi_{\theta_{\mathrm{old}}}(\cdot \mid q)}
\Bigg[ 
\frac{1}{\sum_{i=1}^{G}\lvert \tilde{o}^{i}\rvert}
\sum_{i=1}^{G}\sum_{t=1}^{\lvert \tilde{o}^{i}\rvert}
\\
\min\!\Big(
r^{i}_{t}(\theta)\,\hat{A}^{i}_{t}&,
\operatorname{clip}\!\big(r^{i}_{t}(\theta),\,1-\epsilon,\,1+\epsilon\big)\,
\hat{A}^{i}_{t}
\Big) 
- \beta\, \mathbb{D}_{\text{KL}}(\pi \| \pi_{\text{ref}})
\Bigg],\\[2pt]
\end{aligned}
\end{equation}
where each group contains $G$ rollouts $\{\tilde{o}^{i}\}_{i=1}^G$ sampled from the same query $q$. The advantage $\hat{A}_t^i$ is derived by normalizing rewards $R^{i}$ within a group of $G$ sampled responses $(\tilde{o}^{1}, \ldots, \tilde{o}^{G})$ for each $\tilde{o}^i$:

\begin{equation}
\label{eq:grpo-adv}
\begin{aligned}
\hat{A}_t^i
&= \frac{\,R^{i}-\operatorname{mean}\!\big(\{R^{i}\}_{i=1}^{G}\big)\,}
        {\operatorname{std}\!\big(\{R^{i}\}_{i=1}^{G}\big)} .\\[4pt]
\end{aligned}
\end{equation}

\textbf{Dynamic Sampling Policy Optimization (DAPO).}  
DAPO~\cite{2025dapo} extends GRPO via clip-higher, dynamic sampling, token-Level Policy Gradient Loss, and overlong reward shaping techiniques. 



\textbf{Entropy-aware RLVR.} 
Recent studies have highlighted the importance of token-level entropy in reinforcement learning with verifiable rewards. 
Recent works in math reasoning tasks~\cite{2025beyond,2025stabilizing,2025entropy_review} have shown that high-entropy minority tokens act as critical decision points that determine the trajectory of downstream reasoning. In contrast, low-entropy tokens typically correspond to routine, highly predictable components of the solution, contributing little to downstream reasoning outcome. 
These findings suggest that RL methods should focus gradient updates on high-entropy tokens, rather than uniformly updating all tokens.
Inspired by these findings, we examine entropy patterns in RTL generation and design EARL to exploit high-entropy tokens in Verilog code.



\begin{figure}[t]
    \centering
    \begin{subfigure}[t]{0.48\columnwidth}
        \centering
        \includegraphics[width=1.1\linewidth]{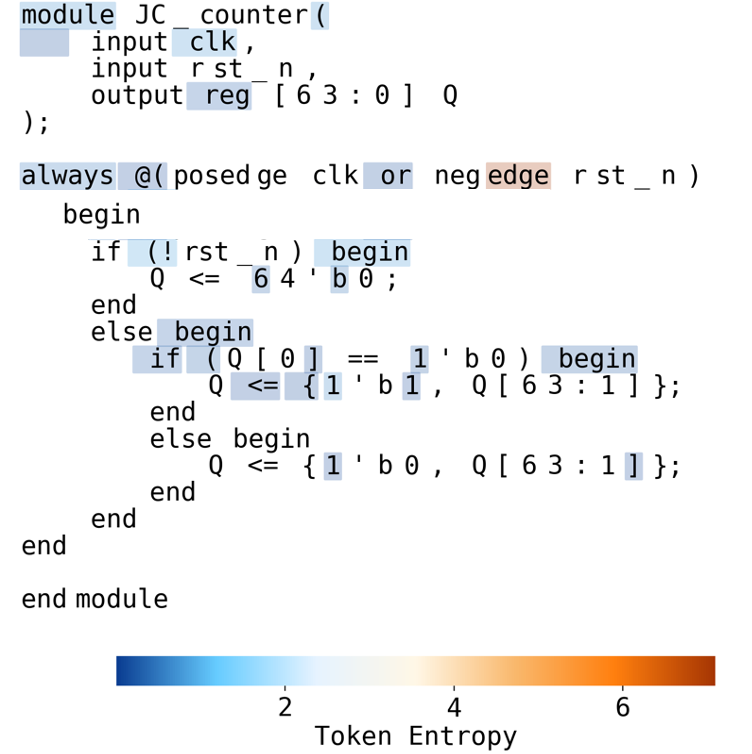}
        \caption{JC\_counter. 
        }
        \label{fig:entropy-counter}
    \end{subfigure}
    \hfill
    \begin{subfigure}[t]{0.48\columnwidth}
        \centering
        \includegraphics[width=1.1\linewidth]{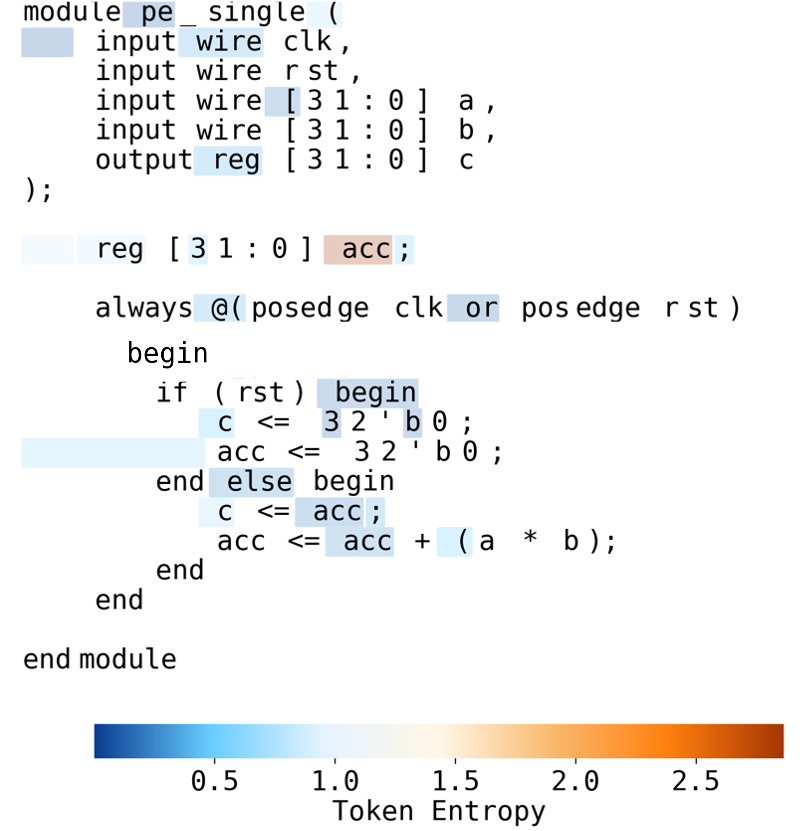}
        \caption{PE\_single.
        }
        \label{fig:entropy-pe}
    \end{subfigure}

    \vspace{-10pt}
    \caption{
        Token-level entropy visualizations for two Verilog modules. 
        Each token is shaded according to its positional entropy, computed over the model’s distribution at that generation step.
        Only high-entropy tokens (above the 80th percentile within each response) are visually highlighted.
    }
    \label{fig:entropy_heat}
    \vspace{-10pt}
\end{figure}

\begin{figure*}[tp!]
\centerline{
\includegraphics[width=\linewidth]{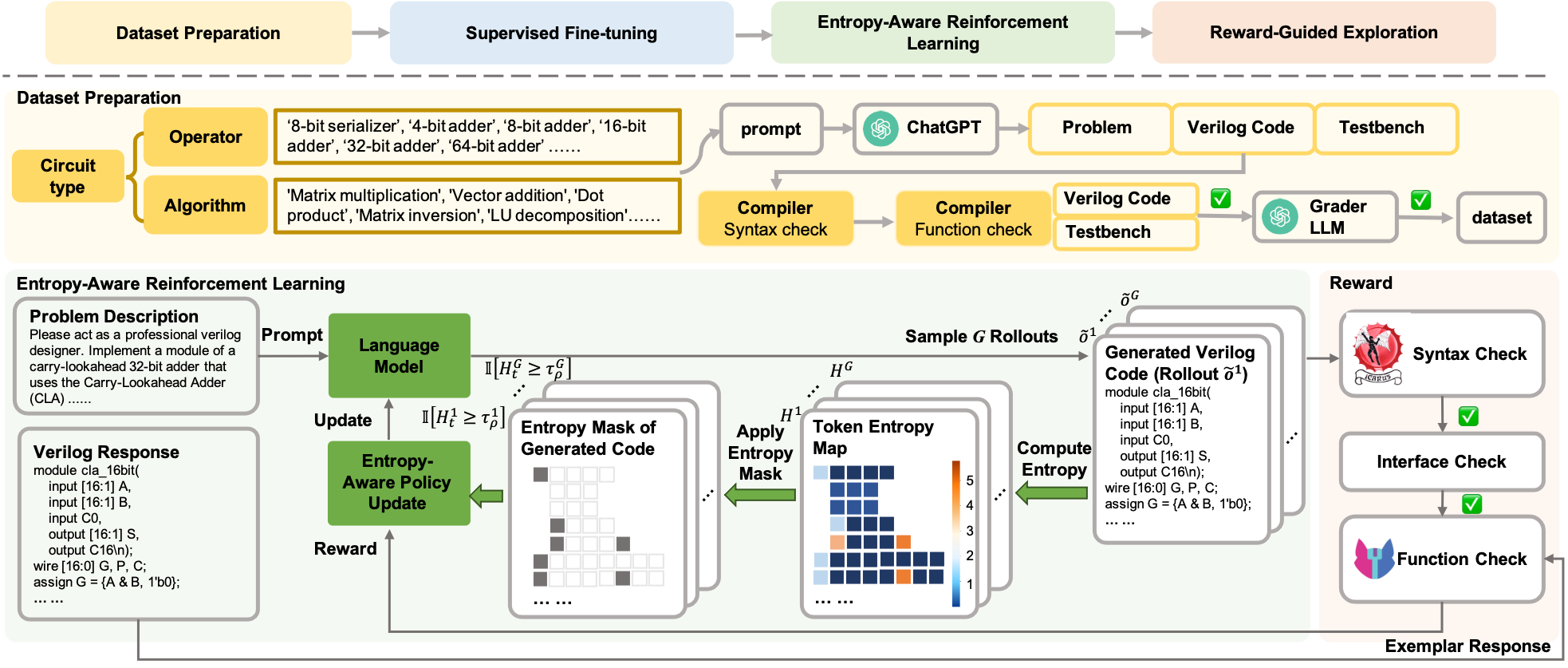}
}
\vspace{-8pt}
\caption{Overview of the EARL framework for Verilog generation. 
The top panel presents the end-to-end pipeline.
The bottom panel details the three core components: dataset preparation (yellow), entropy-aware reinforcement learning (green), and reward-guided exploration (red).
}
\label{fig:framework}
\vspace{-10pt}
\end{figure*}

\section{Token-Level Entropy in RTL Code Generation}\label{sec:entropy}

While prior studies have examined entropy in chain-of-thought reasoning~\cite{2025entropy_review,2025beyond,2025stabilizing}, 
we analyze entropy in RTL code generation at token level.
Specifically, the entropy of the $t$-th token $H_t$ is defined as:
\begin{align}
H_t &= -\sum_{v=1}^{V} p_{t,v}\,\log p_{t,v},\label{eq:entropy}\\
\text{where} \qquad \mathbf{p}_t &= ({p_{t, 1}, ..., p_{t, V}}) \nonumber= \mathrm{Softmax}\!\left(\frac{\mathbf{z}^{\,t}}{T}\right).\nonumber
\end{align}
Here 
$V$ is the vocabulary size, $\mathbf{z}^{\,t}\in\mathbb{R}^V$ is the pre-softmax logits, and $T$ is the decoding temperature. 
Note that entropy is computed per position, not per token type.
For example, the same lexical token (e.g., \texttt{begin}) may appear with low entropy in a deterministic 
context or high entropy in an ambiguous structural context. 
Thus, entropy reflects uncertainty at the generation step, not an inherent uncertainty of a token type.
This metric is used throughout our analysis and later in the EARL training objective.

To analyze entropy behavior in RTL tasks, we begin with concrete examples.
We compute entropy using a DeepSeek-7B coder model that we fine-tuned on our curated specification–implementation dataset. All rollouts use temperature at temperature $T=1.0$. For each generated token $t$, we record its entropy $H_t$ following the definition in Eq.~\eqref{eq:entropy}.
Figure~\ref{fig:entropy_heat} visualizes token-level entropy for two RTL modules from the RTLLM v1.1 benchmark~\cite{rtllm}. For each generated response, only
tokens whose entropy exceeds the 80th percentile of that response are highlighted; low-entropy tokens are shown without background shading.
For instance, in the JC\_counter example (Figure~\ref{fig:entropy-counter}), control-structure keywords (e.g., \texttt{if},
\texttt{begin}) show higher entropy, purely syntactic or declarative tokens (e.g., \texttt{end}, \texttt{endmodule}, numeric literals) exhibit near-zero entropy.
Similarly, in the PE\_single module (Figure~\ref{fig:entropy-pe}), the accumulation logic (e.g., \texttt{acc}) exhibits elevated uncertainty due to its arithmetic and dependency structure.

Beyond these qualitative examples, we analyze entropy statistically over more than $2.1\times 10^5$ generated tokens produced by the same fine-tuned model. Figure~\ref{fig:entropy_hist} presents the overall entropy distribution. The distribution is heavily skewed: over 80\% of tokens fall below $0.15$, while only a small minority exceed $0.6$, forming a distinct heavy tail of high-uncertainty positions.
To identify what these extremes correspond to, Figure~\ref{fig:wordcloud} shows the top-100 lowest-entropy tokens and top-100 highest-entropy tokens with frequency $\geq 10$.  
Two consistent patterns emerge:



(i) {Deterministic bulk vs. uncertain critical tokens.}  
Low-entropy positions correspond almost exclusively to rigid, grammar-driven Verilog constructs such as module headers, port lists, and white space.  
High-entropy positions, in contrast, occur at tokens that govern control or timing behavior—e.g., sensitivity triggers, conditional guards, and data‑path operations—reflecting genuine semantic uncertainty rather than lexical variability.

(ii) {Entropy concentrates at structural junctions.}  
As shown in Figure~\ref{fig:entropy_heat}, the heatmap shows that high-entropy regions appear at sensitivity lists and conditional branches, precisely where module behavior and timing diverge.  
This parallels the notion of “forking tokens”~\cite{2025beyond}, but here the semantic forks represent hardware control flow and signal behavior, not natural language discourse.

These observations confirm that entropy naturally separates deterministic syntax from functional decision points in RTL code.  
This motivates selective updates instead of uniform updates across the entire sequence.
\section{Entropy-Aware RL Framework}

\subsection{Overview of EARL}

To bridge the gap between natural language specifications and functionally correct Verilog, we propose \textbf{EARL}, an \emph{Entropy-Aware Reinforcement Learning} framework tailored for RTL generation. Figure~\ref{fig:framework} illustrates the complete pipeline. EARL addresses key limitations in prior LLM-for-Verilog approaches, including the inability of supervised fine-tuning alone to capture functional correctness signals, and the inefficiency of uniform reinforcement learning updates across long, structured RTL sequences.

The pipeline consists of four tightly coupled stages.
\textbf{(1) Dataset Preparation.} We curate a synthetic dataset of natural-language specifications, Verilog implementations, and corresponding testbenches. This dataset provides functionally verified training triples that ensure syntactic validity and semantic coverage across diverse design tasks.
\textbf{(2) Supervised Fine-tuning.} A pre-trained code LLM is fine-tuned on the curated dataset to establish strong syntactic fluency and structural priors. This initialization equips the model with the ability to reliably generate RTL modules with correct grammar and interface formats.
\textbf{(3) Entropy-Aware Reinforcement Learning.} Building on SFT initialization, we introduce EARL, which integrates verifiable compiler/testbench signals with entropy-guided selective updates. Unlike existing RLVR methods that propagate gradients uniformly across tokens, EARL emphasizes high-entropy tokens that disproportionately govern module structure, control flow, and timing behavior, while preserving conservative updates on deterministic syntax tokens.
\textbf{(4) Reward-Guided Exploration.} To align generation with downstream hardware requirements, we design a cascaded multi-signal reward incorporating syntax validity, interface consistency, and functional equivalence checking. The hierarchical reward mirrors industrial verification pipelines, ensuring that the model’s exploration budget is allocated to meaningful improvements rather than syntactic noise.

By integrating dataset curation, supervised initialization, and entropy-aware RL with verifiable signals, EARL achieves robust one-pass RTL generation. The framework simultaneously improves functional correctness, stabilizes RL training on long sequences, and provides a general recipe for incorporating entropy-driven optimization into RLVR pipelines.

The following subsections describe each component of the framework in detail,
including the formulation of entropy-aware RL, the design of verifiable rewards,
and the dataset construction pipeline.

\begin{table*}[thp!]
\centering
\caption{Comparative analysis of Verilog code generation performance. \textbf{Bold} indicates best result. \underline{Underline} indicates the second best result.
}
\label{tab:main_results}
\vspace{-8pt}

\setlength{\tabcolsep}{3pt} 
\scriptsize
\resizebox{\textwidth}{!}{
\begin{tabular}{llc|cccc|cc}
\toprule
\textbf{Category} & \textbf{Method} & \textbf{Params.} &
\multicolumn{2}{c|}{\textbf{VerilogEval-Machine}} & 
\multicolumn{2}{c|}{\textbf{VerilogEval-Human}} &
\multicolumn{2}{c}{\textbf{RTLLM}}  \\
& & &
pass@1 & pass@5 &
pass@1 & pass@5 &
Syn. pass@5 & Func. pass@5  \\
\midrule

\multirow{5}{*}{\textbf{Base Model}} 
& Qwen2.5-Coder & 1.5B & 25.6 & 40.8 & 8.3 & 17.9  & 75.1 & 19.2 \\
& Qwen2.5-Coder   & 3B   & 48.4 & 58.9 & 21.3 & 32.7  & 82.9 & 27.3  \\
& Qwen2.5-Coder   & 7B   & 52.7 & 69.7 & 23.9 & 41.1 & 86.2 & 31.0\\
& DeepSeek-Coder       & 6.7B &  8.8 & 34.3 & 4.9  & 19.3 & 89.7 & 41.2  \\
& CodeLlama            & 7B   & 26.1 & 49.1 & 18.8 & 28.6 & 15.2 & 9.3 \\
& CodeQwen-7B-Chat     & 7B   & 29.1 & 61.9 & 14.8 & 36.8 & 16.1  & 11.0 \\

\midrule

\multirow{9}{*}{\textbf{Supervised Fine Tuning}} 
& VerilogEval \cite{verilogeval}           & 16B  & 46.2 & 67.3 & 28.8 & 45.9 & N/A & N/A \\
& ChipNeMo   \cite{chipnemo}            & 70B  & 53.8 & N/A  & 27.6 & N/A  & 28.0  & 20.7 \\
& RTLLLM \cite{rtllm}                & 13B  & 65.3 & 77.2 & 43.7 & 51.8 & 52.6  & 40.7 \\
& RTLCoder \cite{xie2024rtlcoder}      & 6.7B   & 37.2 & 64.9 & 16.9 & 35.7  & 89.1 & 51.7 \\
& OriGen \cite{pku2024origen}      & 7B   & 43.1 & 55.1 & 30.8 & 31.9 & \underline{96.5}  & {51.7} \\
& VeriGen \cite{nyu2024}      & 16B   & 44.0 & 52.6 & 30.3 & 43.9 & N/A  & N/A \\
& HaVen \cite{haven2025}      & 6.7B   & 66.1 & 81.1 & 43.1 & 55.1 & 81.4  & 38.6 \\
& BetterV \cite{2024betterv}      & 7B   & 68.1 & 79.4 & 46.1 & 53.7 &  N/A & N/A\\
& AutoVCoder \cite{2024autovcoder}      & 7B  & 68.7 & 79.9 & \underline{48.5} & 55.9 &  \textbf{100}  & 51.7 \\

\midrule

\multirow{2}{*}{\textbf{Reinforcement Learning}} 
& VeriReason \cite{rl2025verireason}           & 7B  & \underline{69.8} & \underline{83.1} & {47.9} & \underline{58.4} & N/A  & N/A \\
& VeriSeek   \cite{2025rl_1}            & 6.7B  & 61.6 & 76.9  & N/A & N/A  & 94.8 & \underline{54.2} \\
\midrule

\multicolumn{2}{c}{\textbf{EARL (ours)}}
& 7B & {\textbf{72.9}} & {\textbf{83.9}} & {\textbf{49.6}} & {\textbf{60.2}}  & {\textbf{100}} & {\textbf{68.9}} \\



\bottomrule
\end{tabular}
}
\end{table*}

\subsection{Entropy-Aware RL}




Our entropy study in Section \ref{sec:entropy} shows that only a minority of high-entropy tokens drive effective learning. 
However, existing RLVR pipelines apply uniform gradient updates across all tokens, which dilutes learning signals and wastes gradient budget on low-impact positions.
To address this limitation, we introduce an 
entropy-gated policy optimization method 
that selectively emphasizes high-uncertainty tokens during policy optimization (see green block in Figure~\ref{fig:framework}).
For each rollout $\tilde{o}^i = (o_1^i, \dots, o_{T_i}^i)$, we compute the per-token entropy $H_t^i$ as defined in Eq.~\eqref{eq:entropy}.
To mitigate cross-prompt entropy scale variation, we adopt response-level quantile masking: a token $t$ is updated only if $H_t^i \ge \tau_\rho^i$, where $\tau_\rho^i = \operatorname{Quantile}(\{H_t^i\}_{t=1}^{T_i}, \rho)$ for a chosen threshold $\rho \in (0, 1)$.
Specifically, tokens whose entropy exceeds $\tau_{\rho}^{i}$ receive full gradient updates, while low-entropy tokens are down-weighted to zero, preserving their stable distributions learned by SFT. 
Formally, EARL extends a DAPO-style objective with entropy gating
:
\begin{equation}
\label{eq:earl}
\begin{aligned}
J_{\mathrm{EARL}}(\theta) &= \mathbb{E}_{(q, a) \sim \mathcal{D}, \{\tilde{o}^i\}_{i=1}^{G}  \sim \pi_{\text{old}}(\cdot|q)} 
\Bigg[
\frac{1}{\sum_{i=1}^{G} |\tilde{o}^i|} 
 \sum_{i=1}^{G} \sum_{t=1}^{|\tilde{o}^i|}
 \\ 
 \mathbb{I} [H_t^i \ge {\tau}_{\rho}^{i}]
 &\Big(
   \min \big( r_t^i(\theta)  \hat{A}_t^i, 
              \text{clip}\!\left(r_t^i(\theta), 1 - \varepsilon_{low}, 
              1 + \varepsilon_{high}\right) 
              \hat{A}_t^i \big) \\
- \beta\, \mathbb{D}_{\text{KL}}(\pi& \| \pi_{\text{ref}})
 \Big)
\Bigg],
\quad \text{s.t.}\quad 
0 < \left|\left\{ i \mid R^i > 0 \right\} \right| < G,
\end{aligned}
\end{equation}
where, the token-level advantage $\hat{A}_t^i$ follows Eq.~\eqref{eq:grpo-adv}. 
The indicator $\mathbb{I} [H_t^i \ge {\tau}_{\rho}^{i}]$ is the entropy mask which differentiates the optimization strength across token type.
The constraint enforces diversity within each group, preventing degenerate all-pass or all-fail groups, which would invalidate the group-normalized advantage.

Note that this design is optimizer-agnostic. EARL can be applied to policy gradient objectives without modifying their core update rules. In our experiments, we instantiate EARL with both PPO and DAPO. By concentrating gradient updates on uncertain yet functionally critical tokens, EARL achieves sharper credit assignment, reduces wasted updates on deterministic syntax, and improves the stability of RLVR training for long, structured RTL sequences.

\subsection{Reward Design}
A central component of EARL is the design of verifiable rewards that align generation with downstream RTL correctness.
Inspired by industrial EDA flows, we construct a cascaded reward hierarchy (see red block of Figure~\ref{fig:framework}), which mirrors the standard compilation–simulation–verification pipeline.
Given a natural language specification $q$, the model generates a rollout $\tilde{o}$ sampled from its current policy. To assign a reward, we evaluate $\tilde{o}$ through a cascaded three-stage verification process:
First, we check syntax validity by compiling the candidate with \texttt{iverilog}. Programs that fail syntax checks are immediately assigned zero reward, preventing wasted computation on invalid code.
If syntax passes, we proceed to evaluate interface consistency. We examine whether the module name and port declarations match the specification. Partial matches receive intermediate reward, while fully aligned interfaces receive higher credit.
Finally, we assess functional equivalence. We invoke the \texttt{eqy} engine from Yosys to compare the output against a reference implementation. Passing candidates receive maximal reward, while near-misses are graded slightly lower to encourage exploration without reward hacking. 



\subsection{Dataset Preparation}

Training EARL requires functionally verified data to bootstrap supervised initialization and support reward evaluation during RL.
To this end, we construct a synthetic dataset of specification–code– testbench triplets (see yellow block of Figure~\ref{fig:framework}).
The dataset generation pipeline follows two steps.
The first step is synthetic generation. 
We prompt GPT-4 to generate diverse problem descriptions covering arithmetic operators, sequential logic, control modules, and algorithmic circuits. Each problem description is paired with a Verilog implementation and an executable testbench. Inspired by prior work~\cite{2024autovcoder,2024betterv}, we diversify the prompts along multiple axes, including difficulty level, circuit type, and specification style.
The second step involves compiler-driven filtering.
To guarantee correctness, we apply a three-stage validation pipeline to all generated samples.
First, syntax validity is verified using \texttt{iverilog}.
Second, functional correctness is checked by executing the paired testbench.
Finally, we employ a grading LLM to provide an additional layer of semantic evaluation on functional behavior.
Only samples that pass all three stages are retained in the dataset.

\section{Experiments}

\subsection{Experiment setup}

\begin{figure*}[t!]
\centerline{
\includegraphics[width=0.85\linewidth]{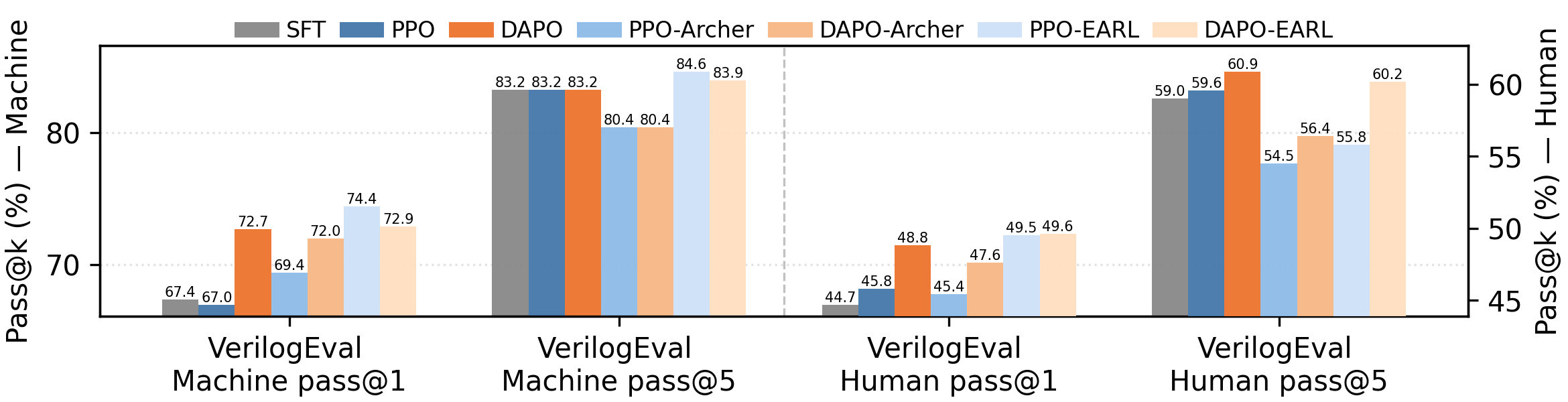}
}
\vspace{-6pt}
\caption{Ablation study on RL algorithms and entropy mechanisms. 
    }
\label{fig:entropy_tech}
\end{figure*}

\textbf{Models and Training.} 
We adopt \texttt{DeepSeek-Coder-7B} as the base model. 
The SFT stage is conducted on our curated dataset for 3 epochs using a cosine learning rate schedule with a peak learning rate of $5\times10^{-5}$ and 15 warm-up steps. 
For reinforcement learning, we apply the proposed EARL algorithm instantiated with DAPO, using group sampling ($G=6$ rollouts per prompt) and a learning rate of $1\times10^{-6}$. 
All experiments are conducted on 4 NVIDIA A100 80GB GPUs with a global batch size of 128 and temperature 1.0 during inference.

\textbf{Benchmarks.} We evaluate on three standard RTL code generation benchmarks.
 \textbf{VerilogEval v1}~\cite{verilogeval} includes 143 machine-generated problems (\texttt{VerilogEval-Machine}) and 156 expert-written tasks (\texttt{VerilogEval-Human}), designed to test structural and semantic generation capabilities.
\textbf{RTLLM v1.1}~\cite{rtllm} contains 29 diverse RTL design prompts focused on real-world functionality.

\textbf{Metrics.} Following prior works~\cite{chipnemo, rtllm}, we use the pass@k metric to evaluate the probability of generating a correct solution within $k$ attempts, where $k \in \{1, 5\}$. For each prompt, we generate $n = 5$ completions and compute:
\[
\text{pass@k} := \mathbb{E}\left[1 - \frac{{\binom{n-c}{k}}}{\binom{n}{k}}\right],
\]
where $c$ is the number of successful completions. 

\subsection{Comparison with Competing Methods}

We evaluate the syntactic and functional correctness of {EARL} against a wide range of baseline and state-of-the-art methods, including SFT and RLVR approaches. Note that for completeness, we also report results of prompt-engineering based methods, e.g., Origen, HaVen. Their pass-k metrics are obtained from our own evaluation under the same experimental setup. 
Results in Table~\ref{tab:main_results} demonstrate that EARL achieves the best performance across all benchmarks and metrics.
On {VerilogEval-Machine}, EARL attains 72.9\% pass@1 and 83.9\% pass@5, surpassing the strongest RL baseline VeriReason by 3.1\% and 0.8\%, respectively, and outperforming the best SFT model by 4.2\% in pass@1. 
On VerilogEval-Human, which contains expert-written tasks more aligned with real-world HDL practices, EARL achieves 49.6\% pass@1 and 60.2\% pass@5, both the highest among all methods. 
On {RTLLM v1.1}, EARL obtains perfect syntax pass@5 (100\%) and the highest functional correctness (68.9\%), outperforming the best RL method by 14.7\% and the best SFT method by 17.2\%. 
These consistent improvements demonstrate that EARL establishes a new state-of-the-art in Verilog code generation, achieving reliable one-pass correctness while remaining parameter- and data-efficient.

\begin{table}[t]
\centering
\caption{Comparative analysis of threshold on Verilog code generation performance. \textbf{Bold} indicates best result. \underline{Underline} indicates the second best result.
}
\label{tab:threshold}
\vspace{-8pt}

\setlength{\tabcolsep}{3pt} 
\scriptsize
\resizebox{\linewidth}{!}{
\begin{tabular}{l|cccc|cc}
\toprule
\textbf{Quantile}  &
\multicolumn{2}{c|}{\textbf{VerilogEval-Machine}} & 
\multicolumn{2}{c|}{\textbf{VerilogEval-Human}} 
&
\multicolumn{2}{c}{\textbf{RTLLM}}  \\
& 
pass@1 & pass@5 &
pass@1 & pass@5 &
Syn. & Func. \\
\midrule
{0.0}
 & \underline{72.7} & {\underline{83.2}} & \underline{48.8} & {\textbf{60.9}} & {94.8} & {62.0}\\
{0.2}
 & {70.5} & {79.7} & {47.3} & {58.4} & \textbf{100} & {\textbf{72.4}}\\
{0.4}
 & {69.4} & {80.4} & {47.6} & {56.4} & {96.1} & {65.5}\\
 {0.6}
 & {67.8} & {77.6} & {42.3} & {53.2} & {95.1} & {65.5}\\
 {0.8}
 & {\textbf{72.9}} & {\textbf{83.9}} & {\textbf{49.6}} & {\underline{60.2}} & {\textbf{100}} & {\underline{68.9}}\\
{0.9}
 & {71.0} & {79.7} & {45.6} & {56.4} & \underline{96.4} & {65.5}\\

\bottomrule
\end{tabular}
}
\vspace{-15pt}
\end{table}

\subsection{Ablation Study}

\subsubsection{Effect of RL Algorithms and Entropy Mechanisms.}

To better understand the effect of our entropy-aware reinforcement learning framework, we conduct an ablation study comparing different RL algorithms (PPO and DAPO) and different entropy mechanisms (Archer~\cite{2025stabilizing} and the proposed EARL). The results are summarized in Fig.~\ref{fig:entropy_tech}.

We first compare PPO and DAPO without entropy masking. DAPO consistently outperforms PPO, confirming the benefits of group-based dynamic sampling in Verilog code generation. Next, we introduce entropy-aware variants. Archer applies entropy weighting, while our EARL design incorporates an entropy mask to selectively emphasize high-entropy tokens. Both methods improve over plain PPO/DAPO, but EARL achieves larger gains, particularly on VerilogEval-Machine pass@1 and VerilogEval-Human pass@5.
Overall, DAPO-EARL achieves the best performance across all benchmarks, with improvements of +5.2\% in pass@1 over the SFT baseline.

\subsubsection{Entropy Quantile.}
We further ablate the effect of the entropy quantile $\rho$ used to construct the response-level mask.
Results in Table~\ref{tab:threshold} show that performance peaks around $\rho=0.8$.
Lower thresholds (e.g., 0.2) admit excessive noise, while higher thresholds (e.g., 1.0) exclude too many decision-critical tokens. 
This validates that selectively emphasizing a minority of high-entropy tokens yields the best trade-off between exploration and stability.

\section{Conclusion}

In this work, we addressed the challenge of reliable RTL generation with large language models.
Motivated by our entropy analysis, we proposed EARL, an entropy-aware reinforcement learning framework that integrates supervised initialization, verifiable reward signals, and selective policy updates. 
EARL emphasizes high-uncertainty tokens that govern module structure, control flow, and signal connections, while preserving stable distributions on deterministic syntax tokens.
Extensive evaluation on standard benchmarks demonstrates that EARL consistently outperforms existing state-of-the-art methods.

\bibliographystyle{ACM-Reference-Format}
\bibliography{sample-base}

\end{document}